\pdfoutput=1

\documentclass[11pt]{article}

\usepackage{acl}

\usepackage{times}
\usepackage{latexsym}
\usepackage{enumerate}
\usepackage{caption}
\usepackage{subcaption}
\usepackage{graphicx}

\usepackage[T1]{fontenc}

\usepackage[utf8]{inputenc}

\usepackage{microtype}

\makeatletter
\newcommand{\printfnsymbol}[1]{%
  \textsuperscript{\@fnsymbol{#1}}%
}
\makeatother

\title{\textit{Not always about you:} Prioritizing community needs \\ when developing endangered language technology}

\author{Zoey Liu $^*$ \\
  Boston College \\
  \texttt{zoey.liu@bc.edu} \\\And
  Crystal Richardson (Karuk) $^*$ \\
  University of California, Davis \\
  \texttt{cjrichardson@ucdavis.edu} \\\AND
  Richard Hatcher Jr \\
  University at Buffalo \\
  \texttt{rjhatche@buffalo.edu} \\\And
  Emily Prud'hommeaux \\
  Boston College \\
  \texttt{prudhome@bc.edu} \\}

\begin{document}
\maketitle

\def\thefootnote{*}\footnotetext{Equal contribution.}\def\thefootnote{\arabic{footnote}}


\begin{abstract}
 Languages are classified as \textbf{low-resource} when they lack the quantity of data necessary for training statistical and machine learning tools and models. Causes of resource scarcity vary but can include poor access to technology for developing these resources, a relatively small population of speakers, or a lack of urgency for collecting such resources in bilingual populations where the second language is high-resource. As a result, the languages described as low-resource in the literature are as different as Finnish on the one hand, with millions of speakers using it in every imaginable domain, and Seneca, with only a small-handful of fluent speakers using the language primarily in a restricted domain. While issues stemming from the lack of resources necessary to train models unite this disparate group of languages, many other issues cut across the divide between widely-spoken low-resource languages and endangered languages. In this position paper, we discuss the unique technological, cultural, practical, and ethical challenges that researchers and indigenous speech community members face when working together to develop language technology to support endangered language documentation and revitalization. We report the perspectives of language teachers, Master Speakers and elders from indigenous communities, as well as the point of view of academics.
 We describe an ongoing fruitful collaboration and make recommendations for future partnerships between academic researchers and language community stakeholders.

\end{abstract}

\section{A thought experiment}
\label{thought}

Say that we have three speech communities, which we will refer to as Elephant, Ocelot, and Coyote.
Each community has their own language, which is commonly characterized by language technology researchers as ``low-resource'' or ``under-resourced''.
Elephant's language is spoken by around 10 million first language (L1) speakers and millions more second language (L2) speakers, and it has a standard widely accepted orthography. 
Ethnologue~\citep{lewis2015ethnologue,campbell2008ethnologue} classifies this language as ``used in education, work, mass media, and government at the national level.''
Ocelot's language has about 500,000 L1 speakers and three different orthographic representations.
This language is noted in Ethnologue as ``in vigorous use, with literature in a standardized form being used by some though not yet widespread or sustainable'', and is described as an ``indigenous'' language.
In comparison, the language of Coyote has only a handful of speakers that could be considered fluent L1 speakers, most of whom are elders.
The language currently lacks a standard orthography, and in Ethnologue it is classified as ``endangered''~\citep{meek2012we}.

Elephant wants to develop spoken language technology primarily to support the use of speech-enabled applications and smartphone features.
Ocelot and Coyote also need spoken language technology but their focus is often on using these technologies to support use of the language by the community through language preservation, documentation, and instruction. Creating chatbots or enabling hands-free cellphone use, while appealing, might not currently be a priority for Ocelot and Coyote.

\subsection{Collecting training data}
With support from local universities and the government, Elephant has started a project to build an automatic speech recognition (ASR) system.
Thus far it has collected around 60 hours of audio data gathered from radio stations as well as recordings produced in a studio.
To transcribe the recordings in the common orthography, Elephant resorted to online crowd-sourcing platforms.
It has gathered additional texts scraped from the web and from newspapers and books; the texts also followed the standard orthography of the language, totaling several million tokens.
All of the preparations required financial resources and time, but they were completed within the course of less than a year.
Elephant's language technology development work has received interest and accordingly collaboration offers from potential industry partners, including a company that makes a popular language learning application.
In return the company has asked for the language data from the project either to be made public or to be owned by the company. Elephant decided to cooperate.

Ocelot wanted to try ASR as well, though it did not know where to start at first and did not at the moment have government support. On the other hand, Ocelot has connections with a professor in Indigenous Studies who has advanced research knowledge of the language and is trusted by the language community. 
Over the years the professor has collected around 100 hours of recordings produced by various community members.
Though there was not yet a standard orthography for the language, some agreement was reached among the community members. 

Unlike Elephant, Ocelot was not able to use crowd-sourcing platforms since there are not that many people who have literacy in the language.
Fortunately, the professor was able to recruit students from the local university who are L1 speakers of the language to transcribe audio data.
They also managed to gather digitized texts of $\sim$250,000 tokens from available websites and converted them to the same orthography used for the transcriptions.
All of these preparations were finished in a few years.
When the same company that reached out to Elephant made the same offer, Ocelot said she would think about it.

Things were quite different for Coyote. She did not have support from the government or any connections with universities. A professor used to work with elders from the community and collected dozens of hours of recordings from fieldwork sessions throughout the years, but at a certain point he stopped giving the audio data back to the community and there were no additional copies. The few recordings made by anthropologists and linguists in the early 20th century were not yet digitized and remain archived outside the community. In other words there was no available data yet to build speech technology.
When the same company offered to make learning applications in exchange for ownership of the language data, Coyote said no because of their prior negative experiences with outside researchers and the government-funded attempted linguicide~\citep{hinton1994flutes,skutnabb2000linguistic} that occurred in their community just a few generations ago. At this point, many members of the Coyote language are reluctant to work with outsiders, even corporations willing to help them.

Coyote decided to be in charge of the documentation process herself. Right now there is one community member who is an L2 learner and is studying computer science as a student at a local university. She is trying to learn about how to build ASR tools. At the same time, she is working with internal researchers from Coyote, some of whom do not have formal training on the linguistic aspects of their language.
Their data collection process, however,  has turned out to face severe challenges. 

First, recordings are gradually obtained from conversations with the elders, who are kindly working for free. In order to be respectful to the elders' schedules and to make sure they get enough rest during the long fieldwork sessions, the audio collections are  ongoing and comparatively much slower than that for the languages of Elephant and Ocelot.

Secondly, the lack of standard orthography for the language creates difficulty in choosing reasonable representations. Although the language has one available grammar book that was written back in the 1920s, there are many words in the recordings that do not appear in the grammar. Additionally, having representative community members come to a consensus on a single orthography requires extensive time and discussion.
Over the decades various linguists and anthropologists have come up with a total of eight different orthographic representations for the language, but having multiple written orthographies has resulted in several possible pronunciations for one word or one single utterance. Thus, a significant amount of deliberation is required in order to ensure that the speakers from Coyote do not have too much difficulty reclaiming words or sentences from written documentation.
Writing is seemingly so common for languages that are widely spoken or studied in the world that people tend to take it for granted; they do not exactly realize the writing is itself a luxury and a form of language technology that is not natural to many oral speech communities~\citep{bird-2020-decolonising,hinton1994flutes,ryon2002cajun,crystal2018}.


Thirdly, the transcribing process is an extremely time-consuming endeavor. Although the graduate student herself is an advanced L2 learner, many of the other learners do not have the same proficiency. In other words, there are very few people from Coyote who are capable of transcribing recordings of their language, commonly referred to as the ``transcription bottleneck"~\citep{zahrer-etal-2020-towards,shi-etal-2021-leveraging}. Therefore the transcriptions have to be cross-checked through consultations with the elders from the Coyote speech community from time to time.
At this pace, over five years, around only 16 hours of recordings for Coyote have been collected and transcribed, among which 6 hours are monolingual narratives and story-telling in the language, while the rest includes a large amount of code-switching with English.
Additional written texts for the language were digitized from the grammar book and an available Bible in the language, yielding around 40,000 tokens.

\subsection{Training the ASR models}
Now each of the three speech communities has some training data in their own language. They are ready to train ASR systems.
Elephant adopted the deep neural network (DNN) from the popular Kaldi toolkit~\citep{povey2011kaldi}. The transcripts of the audio data used as training data and additional written texts were combined to build a language model, which was then applied in the decoding process of the acoustic model. Word error rates (WERs) below 20\% were achieved. Elephant also tried the newer end-to-end neural ASR library, ESPNet~\citep{watanabe2018espnet}, which does not necessarily require language models and has been demonstrated to work well for languages with dozens of hours of audio data (e.g., Yolox{\'o}chitl Mixtec~\citep{shi-etal-2021-leveraging}) or more (e.g. English, Hindi~\citep{khare2021low}); a similarly strong WER was obtained.
Ocelot engaged in the same efforts and she was able to obtain WER numbers comparable to those for Elephant.

In contrast, when applying the same DNN architecture from Kaldi (with different hyperparameters) to her six hours of monolingual audio data, Coyote was able to derive a WER of just under 40\%. Coyote thought that perhaps this had something to do with the language model since it was trained on a very small number of words so she turned to ESPnet, opting for a training scheme without using the language model for decoding. The results were even worse. Similar results were produced using wav2vec-U~\cite{baevski2021unsupervised}. Even the ASR frameworks touted as particularly successful for low-resource languages yielded disappointing results for Coyote's language.

On the other hand, the number 40\% means something different for a community of speakers than it does for the research community. A WER this high might not be that impressive for (academic) researchers, but it may be \textit{good enough} in the meantime; a model trained on the available six hours of audio data can be used to generate transcriptions of new recordings from ongoing fieldwork sessions or untranscribed archival recordings. These transcriptions can then be corrected by L2 speakers, expediting the transcription process and creating new acoustic and textual training data~\citep{prudhommeaux2021}.

For speech communities like Elephant and Ocelot, which are relatively widely-spoken and often benefit from financial support or collaborative bonds with academics and industry partners, it is possible to collect more data, whenever it is needed. For endangered languages like Coyote's, however, it is unlikely that it will ever be possible to gather even dozens of hours of audio data, regardless of financial or time constraints.

\section{Academic perspectives}

\begin{table*}[ht!]
\small
\centering
\begin{tabular}{llll}
\hline
\textbf{Category} & \textbf{Description} & \textbf{Count} & \textbf{Examples}  \\ \hline
Elephant & widely spoken, well supported & 99 (60.7\%) & Bengali, Danish, Igbo, Pashto, Tagalog \\
Ocelot  & fewer speakers, well supported & 39 (23.9\%)  & Faroese, Maori, Quechua, Yiddish \\
Coyote & few speakers, little support  & 25 (15.3\%)  & Bribri, Kodi, Mi'kmaq, Veps, Yine \\    \hline
\end{tabular}
\caption{Number of unique languages named in ACL Anthology abstracts that include the phrases \emph{low resource, under resourced} or \emph{resource constrained language}, organized by low-resource language type. }\label{aclanth}
\end{table*}

As illustrated above, the current research field has coarsely referred both to widely spoken languages lacking an established tradition of natural language processing (NLP) and to endangered indigenous languages as ``low-resource" or ``under-resourced", without acknowledging or mentioning the drastically different conditions of their data availability. 
Recent high-profile work on diversity in language technology included, in one case, just a few sentences encouraging researchers to prioritize endangered languages ~\citep{blasi2021systematic}, and in another case, no discussion of endangered languages at all~\citep{mager-etal-2020-tackling}.
While the language taxonomy based on resource availability proposed in~\citet{joshi-etal-2020-state} is quite impressive, the authors simply grouped all languages that are currently lacking resources into the same category. For instance, the Mixtec language (with different varieties), spoken in Mexico, has around half a million L1 speakers, while the Juruna language in Brazil has less than 300 L1 speakers; yet both were categorized together as ``still ignored in the aspect of language technology".

Roughly 1,050 of the nearly 30,000 abstracts available in the ACL Anthology bibliography contain the token phrases ``low resource'', ``under resourced'', or ``resource constrained language''. Of these abstracts, about half name a specific language that has been assigned an ISO-639 code. Excluding obviously high-resource languages that are mentioned in these abstracts as a point of comparison, as a source language for transfer learning, or as a source or target for machine translation, we are left with 163 unique languages (considering their dialectal variations) that have been characterized in at least one ACL abstract as low-resource. Table~\ref{aclanth} shows the distribution of these 163 languages with similar resource conditions/external support as Elephant, Ocelot, or Coyote respectively, along with a few examples of each language category. We see Elephant languages far outnumber both Ocelots and Coyotes, with Coyotes representing only 15\% of the languages identified as low-resource.

There is great variability both in the degree and in the nature of the challenges that arise when developing language technologies for languages with scarce training resources. It is crucial that NLP researchers actively acknowledge this variability and avoid giving the impression that models or architectures developed for Elephant will be suitable for Ocelot or Coyote, or vice versa. 
We can begin by distinguishing languages classified as ``endangered'' from those that are not, and provide case-by-case detailed descriptions of the speaker population size and language data availability for the language being investigated.
Contingent on that, it is only recently that the academic community has started holding workshops devoted to endangered and indigenous languages, such as ComputEL~\citep{computel-2021-use}, held four times since 2014, and AmericasNLP~\citep{americasnlp-2021-natural}, which took place for the first time in 2021.
Work published in these and other venues has included research on several NLP tasks that pertain to language documentation and reclamation for endangered languages, from morphological segmentation~\citep{liu-etal-2021-morphological,kann-etal-2018-fortification}, finite-state morphological analyzers~\citep{lane-bird-2020-bootstrapping,lachler-etal-2018-modeling}, 
to machine translation~\cite{zhang-etal-2020-chren,bird-chiang-2012-machine} and ASR~\citep{thai2020fully,ethan2021,shi-etal-2021-highland}.

That being said, most of the work has focused on technology development, with relatively little regard for the ways in which the development of language technology for endangered languages might be different from that for languages with few existing resources but a much larger numbers of speakers.
Discussions of whether a proposed language technology would be useful for the workflow of the community's own language documentation efforts, or how it would be combined with the community's revitalization and instructional activities are also noticeably lacking, with a few notable exceptions such as the the verb conjugator, Kawenn{\'o}n:nis, developed for the Ohsweken dialect of Kanyen{'}k{\'e}ha~\footnote{https://kawennonnis.ca/about}~\citep{kazantseva-etal-2018-kawennon}; the online dictionary developed for Hupa~\footnote{http://nalc.ucdavis.edu/hupa/hupa-lexicon.php}, which is currently used for language-related activities in the community; and the Indigenous Languages Technology project at {NRC} {C}anada~\cite{kuhn-etal-2020-indigenous}.

The NLP community has formally recognized the importance of developing technologies for endangered languages, and we have the tools to support work in this area. Now we must try to answer this question: what priorities and considerations should researchers take into account when developing technology for endangered languages?


Often NLP technologies presume that a language has a standardized written form that may act as source or target for various computational tasks (e.g., ASR, machine translation, named-entity recognition). While standardization is typical of languages in most of the W.E.I.R.D societies \cite{henrich2010weirdest}, this is atypical for much of the rest of the world. For many endangered language contexts, the tradition of literacy is very recent, and writing is far less privileged than the oral medium. 


Sociolinguistic research on small languages has identified significant variation, both dialectal and ideolectal~\citep{skilton2017three}, in these contexts as well. Standardization, frequently considered a first task in language documentation and the development of language technology, often tends to run counter to the goals of the speech communities working towards revitalization~\citep{whaley2011some}. For these communities, linguistic variation is not a problem to be solved but an important element of a vital language ecology. 


A major goal in language revitalization involves facilitating the usage of the endangered language in a wider set of contexts and situations than it is being used in currently. This entails careful forethought into how language tools can assist in this broadening of usage. Providing state-of-the-art language technology to a community that is critically involved in training new speakers and developing new contexts for usage may not be the most efficient use of time and resources. Tools that assist in classroom education and in developing new usage situations are likely to be of much more immediate value to groups involved in revitalization.

The number of speakers of a language also impacts the maximum rate at which new data can be collected to add to the resources available for a language. Even for a language with just thousands of speakers, documentation projects can accumulate resources at a pace impossible for a highly-endangered language. In Yolox\'ochitl Mixtec, for instance, with around 5000 speakers, there exists a speech corpus of over 100 hours of running speech \citep{mitra2016automatic}. 
For languages with few speakers capable and comfortable of speaking their language, who are mostly elders and also very occupied with other activities related to the revitalization of the language, the goal of collecting long-duration recordings in the language does not seem feasible or even reasonable.


Additionally, while through the decades certain linguistic academic scholars have responsibly and sensitively built trustworthy and collaborative bonds with indigenous communities~\cite{hale1992endangered} (e.g., Dr. Ken Hale working with Warlpiri~\cite{hale1983warlpiri} and Navajo~\cite{remember}; Dr. David Rood working with Lakhota~\cite{rood1996sketch}; Dr. R. M. W. Dixon working with Australian aboriginal languages~\cite{dixon1970proto}), for many endangered language communities in North America, the attitudes of earlier European-American ``settler" scholars towards indigenous communities and their languages have engendered distrust in the motives and biases of outside ``experts'' \citep{harveyNativeTonguesColonialism2015}. Earlier fieldwork often involved outsider linguists paying speaker consultants to participate in research that was designed and conducted solely by the researcher in what is now referred to as the linguist-focused model \cite{czaykowska2009research}. A more recent trend in linguistics is the movement toward Community-Based Language Research, in which community members collaborate with outsider linguists on the research which they themselves help design. In the development of language technology, providing the speech communities a central role in the design and implementation of language tools may improve the likelihood of the tools' success.

\section{Endangered language teachers perspectives}


\begin{table*}[h!]
\small
    \centering
    \begin{tabular}{l|c|c}
    \hline
   \textbf{Question} & \textbf{Yes} & \textbf{No}  \\
 \hline
 Do you think writing could be a technology? & 91.30\% & 8.70\% \\
 Can you read written form(s) of the language? & 95.65\% & 4.35\% \\
 Do you use written documentation when working on language revitalization? & 100\% & -\\
 Do you think morphological segmentation could be a useful technology? & 86.96\% & 13.04\% \\
 Do you think automatic speech recognition could be a useful technology? & 82.61\% & 17.39\% \\
 Do you think video parsing could be a useful technology? & 82.61\% & 17.39\% \\
 Do you think pedagogical learning applications are a useful technology? & 95.65\% & 4.35\% \\\hline

    \end{tabular}
    \caption{Summary statistics of an informal survey on language documentation and technology of language teachers from four endangered language communities of North America.}
    \label{tab:survey_statistics}
\end{table*}

While language technology is certainly of significant value from an academic's point of view, is it actually useful to stakeholders in endangered language communities?
To learn from community voices, we designed an informal survey (Table~\ref{tab:survey_statistics}) and received responses from a total of 23 language teachers coming from four endangered communities. Among them, five are community-designated Master Speakers~\footnote{Master speakers are indigenous community members who are fluent in their language and have accepted apprentices who study the language with them through the oral tradition~\citep{richardson1993now}.} of the language, two of whom are elders; the rest of the respondents include one young L1 speaker, and either semi-fluent or fluent L2 speakers of their languages.


In the survey, we asked for language teachers' thoughts on whether they would consider writing to be a technology, and additionally, whether they think writing, morphological segmentation~\citep{cotterell-etal-2016-joint}, ASR~\citep{jimerson-prudhommeaux-2018-asr}, video processing, and pedagogical learning applications~\citep{bettinson-bird-2017-developing}, which are all common in the research field of NLP, could be useful for them.

Overall, the majority of community language teachers think all of the five technological applications would be helpful but to different extents.
For example, most think that having written documentation would be valuable for aspects of language teaching, reclamation, and intergenerational transmission of cultural and linguistic knowledge. In the words of one respondent, ``Written documentation is useful because a lot of it is old.  It captures the way speakers talked and created words. The writing can be harvested, so that we reclaim our words and speak in the old style again." However, writing has not been shown to be an acceptable \textit{alternative} to learning under a Master Speaker through the oral tradition, which requires no written resources; and learning from writing alone is ``not necessary or sufficient for restoring linguistic and cultural knowledge".\footnote{We acknowledge that different indigenous speech communities have different perspectives on these matters. Writing alone is extremely useful for speech communities with no Master Speakers left. We note particularly the Breath of Life (BOL) speech communities that reclaim their languages from written documentation; the written documentation helps BOL community researchers become the next generation of Master Speakers. In other words, writing is not the end goal in the language learning process for these indigenous communities.}



With morphological segmentation, most language teachers stated that knowing the morphological structures of words would be informative for them to learn to ``piece together the meaning of phrases".  ``Before I attended language pods I had gone to some community language classes, but living far away a lot of my language learning came through reading the dictionary. I thought that having learned the morphologies of words because of how the dictionary shows those word parts was extremely helpful for me in order to not only learn those words but also be able to figure out how to create other ones without checking the dictionary."

In the case of ASR, some language teachers thought that it would be ``an interesting idea"; they could see themselves ``reviewing the transcripts" and the technology would help with ``sound recognition and pronunciation."; these transcripts, however, would not be as effective as ``listening to the audio". Others expressed strong feelings against ASR, saying that ``To me this is just a way for linguists to secure funding for themselves and their tech project, which takes money and resources away from speech communities.  This kind of thing is not language revitalization, as it doesn't create new speakers.  It generates new texts so they can sit on the shelf of some archive. Not helpful".




Almost all language teachers favor automatic processing of video materials. They mentioned this technology is of great value because ``(it) captures the  authenticity of the Speaker/Apprentice"; others said they would use videos ``to watch the elders' mouths when they speak as well as their facial expressions. Intonations
and body expression really add to conversations. Those things can be lost when just reading, writing or
even listening. Seeing the facial expressions and body language are important to understand the contexts
of how specific words are being used".

At last, regarding pedagogical applications, community members suggested that they could be beneficial given that the applications could ``provide repetition, drill and practice.  This would increase learner confidence in their own adequate exposure to the language (in a non-threatening manner).  It would allow the learner to understand how the language works, so that new speech beyond the app could be developed by the learner". They indicated, however, that for the applications to \textit{work}, people would have to actually ``use them".



The observations from this informal survey might be surprising to researchers, who typically consider language technology to be broadly useful and beneficial to all people.
An awareness that is lacking in the research field, however, is that the purpose of language technologies  and their development process might vary significantly when applied to indigenous and endangered languages. For  many of these languages, the priorities of the speech communities are how to more effectively document, teach, and reclaim their language; how to save the cultural heritage passed down from the elders; and how to let their language have a voice among other widely-spoken or dominant languages.

\section{Elder perspectives}
\label{elder}

To linguists, the Karuk speech community is a critically endangered language community of North California.  To speech community members, the Karuk language is a vital language with approximately 25 speakers, including five Master Speakers and other language teachers, archivists and activists.  The Karuk language community is one which is thriving, though surviving through grass roots revitalization with very little infrastructure at the tribal language program level.

When it comes to modern language revitalization, data cannot and should not be separated from the Master Speakers. Their experience of government policies aimed at linguicide~\citep{hinton1994flutes,skutnabb2000linguistic} and their sense of loss of their mother tongue, as L1 speakers become more and more scarce, are realities that new speakers and field linguists need to acknowledge.  For speakers of the Karuk language, these issues come through in the community's internal documentation of their language. For instance, one Master Speaker of the language explained the loss of speakers when he said: ``(t)hose were all my friends.  That's what I was telling [the nurse]. I said I got a lot on my mind.  I said, I sit here all by myself and I'm thinking about all the people that left me.  I said, it's kinda, you don’t feel good I mean, you know, when you think about them.  You're not supposed to, they're gone.  Xâatik, let it go.  But I just can't help it.  I think about all the funny things we did together, laughing and talking"~\footnote{All quotes in this paper were initially documented by an author of this paper during her fieldwork sessions.}. His words reflect the culture of language loss as it occurs on the human level.  Sometimes this reality stops potential Master Speakers from working with linguists and young speakers eager to reclaim their language/identity.  

It is clear that when working with elderly Master Speakers, methodologies must include space for the elders to vent their grief.  Only after this is attended to, can the language be learned by young language workers striving to create a future where the language is worth more than the losses felt.  The young people and Master Speakers who attend to this re-envisioning of an indigenous future with endangered languages in perpetuity, are the cornerstone of hope and healing~\citep{whalen2016healing,hornberger2014bringing,leonard2011challenging}.

One speaker discusses her view of indigenous second language acquisition (SLA): ``Reading outta them books is OK.  You can probably gather a lot.  Yeah, you guys work hard at it.  I've been trying to get all these people together.  At least once a month all the speakers should work together.  You can just gather one day, all the language speakers, you know?  And start talking the language.  A lot of [speakers] get enough [language] so they can teach... But they're not really getting all of it.  If we're all together, maybe you had something that you wanted to say and you didn't know the words, you could ask somebody.  Somebody would know.  In other words, help one another".  Here we come to understand that linguistic documentation is useful for language reclamation, and writing ``them books'' is a fruitful task~\citep{leonard2017,rigney1999internationalization}.  But one theme that emerges from documentation of elderly Master Speakers of the Karuk language is that writing cannot replace the value of speech communities coming together and speaking their language, continuing the oral tradition.  Written data doesn't save a language; it safeguards knowledge.  The ultimate goal of endangered language communities is to someday house that same knowledge in the hearts and minds of their members.


\section{A case study in bridging the indigenous and academic communities}

What does a healthy and trustworthy workflow look like when bringing together an endangered indigenous language community with the (academic) research community?
After gaining perspectives from academics as well as language teachers and elders from indigenous endangered language communities, here we describe an ongoing workflow devoted to developing a morphological parser for the Cayuga language.

With approximately 50 L1 elder speakers and an ever-growing number of L2 speakers, the Cayuga language fits the description of a highly or critically endangered language. Community-based revitalization projects include an immersion language preschool, adult language courses, as well as teacher-training programs at the local Polytechnic. 


Two authors of this paper are participants in this project,
and both are linguists in academia. Specifically, one of them has years of connections with the Cayuga speech community and advanced research knowledge in the language, while the other has extensive training in computational linguistics but no initial connections with the speech community.

This project started in late September, 2021, and is ongoing.
The overall workflow is simple and straightforward. First, the author known by the Cayuga speech community introduced the other author to the community. They described the general idea for the project and mentioned that if they were to carry out the project, they would begin with words already found in the published grammar. While morphological segmentation is of interest to linguists, and morphological supervision has potential utilities for certain NLP tasks such as dependency parsing~\citep{seeker-cetinoglu-2015-graph} and bilingual word alignment~\citep{eyigoz-etal-2013-simultaneous}, in this case, the main goal was to ask whether community members would find morphological segmentation useful for their own language teaching and documentation.
Community members mentioned that explicitly teaching students various inflectional elements of complex verbs, segmenting them, and in some instances color-coding morphemes have been useful for students to learn verbal arguments.

Second, after securing the go-ahead from community members, the two authors have been meeting almost every week for an hour to discuss progress. The author with extensive research knowledge of the language manually performs morphological segmentation of around 50 words every week. In particular, he provides annotations of both \textit{surface segmentation} and \textit{canonical segmentation}~\citep{cotterell-etal-2016-joint}.
The former is to be later incorporated into the workflow for building ASR systems for the language using recordings already collected; the latter has the objective of gaining a better understanding of the language from a linguistic perspective.
The key difference between these two types of segmentation is that for surface segmentation, the concatenation of the individual morphemes stays true to the initial (orthographic) representation of the word (e.g., \textit{onadowa:dǫh} $\rightarrow$ \textit{on-adowa:d-ǫh}; the word means \textit{they are hunting}), whereas for canonical segmentation, decomposing a word into its component morphemes involves the addition and/or deletion of characters (or phonemes) in order to outline the orthographic or phonological changes during the word formation process (e.g., \textit{onadowa:dǫh} $\rightarrow$ \textit{yodi-adowad-ǫh}).

With the new words annotated every week, the author with a computational background trains segmentation models in an iterative fashion, by combining the words of the current week with those from previous weeks to construct a data set for model training and evaluation. 
Model performance, indexed by F1 score, is recorded weekly.
As of now we have annotations for 262 words. The F1 scores for both surface and canonical segmentation approximate 50\%. 
Our follow-up step is to train models using all these words, then apply them to new data that has not yet been annotated to enhance and accelerate manual annotation. Once the F1 scores reach around 75\%, we plan to report back to the community, inform them about where things are, and discuss details of incorporating our research output with their own language work.

\section{Recommendations for ethical collaborations}

Considering academic output on endangered languages more holistically, we conclude that there are not enough narratives about the process of working with the community. 
Academic (NLP) researchers working on indigenous languages, particularly endangered languages like Karuk and Cayuga that have historically been suppressed, should take the following steps when planning projects, describing their research, and collaborating with stakeholders in the speech communities:
\begin{enumerate}[(1)]
\item Make efforts to actually know the indigenous speech communities and build meaningful bonds with them. For instance, fieldwork researchers, when possible, should try to train young community internal researchers to document the language, if they have aspirations to become language teachers; training them can help the speech communities increase longevity and sustainability. Academic researchers should continue to assist the community that they have worked with when Master Speakers are no longer able to participate in language documentation.  Assistance might involve writing dictionaries that would later be given to the community, or helping heritage language learners with language revitalization and reclamation.

\item Consider that speakers and community researchers should be offered the opportunity to be co-authors on work they made meaningful contribution to, and/or be listed as contributors in appropriate ways.

\item Describe the data collection protocols followed and challenges faced in research output. Be attentive and respectful to indigenous community members' schedules, needs (e.g., elders might need to take medications during fieldwork sessions), and perspectives.
\item Speak clearly about plans for the sharing, archiving, and storing of the data~\citep{rigney2006indigenous}. In particular, make sure to be aware that Master Speakers want at the very least co-use copyright over \textbf{all} data which shall be inherited by their descendants.  In addition, physical copies of all data should be given to Master Speakers, and copies should be submitted to tribal archives or archivists.  The only exception to this rule occurs when Master Speakers ask for their data to be edited before being made public to remove gossip or culturally sensitive material before making copies available.
\item Create language technologies together in consultation with speech communities in order to ensure their usefulness to language programs.  The developed technology needs to be accessible to community language workers.  
    \item Discuss concrete plans for how technology output can be incorporated into the documentation and revitalization work of the speech communities. 
    For instance, a morphological parser needs to visualize morphemes and word construction in such a way as to be a valuable teaching tool for speech community members; ASR systems should not require data extraction or facilitate data ownership by community-external researchers or corporations.

\end{enumerate}

Lastly, each perspective and motivation for indigenous language documentation has value and is worth recording. We hope our work will encourage academics to focus on prioritizing the needs and preferences of endangered speech communities when working with them to develop technologies for their languages.
In particular, academics must keep in mind the relationship many endangered language communities have with their languages. One Master Speaker of the Karuk language captured this nature of this relationship when he said ``(t)he Karuk language is a canoe. It holds all of our baskets, our regalia, our materials, our food. The canoe holds all our practices, songs and stories. It holds all our people and all the Karuk people yet to be born. The canoe carries us all; without it, we can’t get anywhere”~\cite{crystal2018}.




\section*{Acknowledgements}

We would like to thank the speech communities of Karuk, Gayog\underline{o}ho:n\k{o}' (Cayuga), Kanienkeha (Mohawk), and Onödowa’ga:’ (Seneca). With the Karuk speech community, we are especially grateful to the following honored Elders who have passed away: Junie Donahue, Vina Smith, Charlie Thom, and Sonny Davis; we are also grateful to the current Master Speakers of the community who provided us with their invaluable insights: Julian Lang, Phil Albers Jr., Nancy Steele, and Susan Gehr; lastly, we want to thank the dedicated language teachers of the community: Florrine Super, Tamara Alexander, Lulu Alexander, Jason Hockaday. 
Within the Gayog\underline{o}ho:n\k{o}' speech community, we are especially grateful to Gasenneeyoh Crawford, Renae Hill, Amos Keye, and Sose Smith.
This material is based upon work supported by the National Science Foundation under Grant \#2127309 to the Computing Research Association for the CIFellows Project, and Grant \#1761562.
Any opinions, findings, and conclusions or recommendations expressed in this material are those of the author(s) and do not necessarily reflect the views of the National Science Foundation nor the Computing Research Association.

\section{Ethical consideration}

The data from the informal survey described in this paper were collected from language teachers and speakers from four indigenous speech communities of North America. The survey is language-agnostic in the sense that it could be used and expanded to other indigenous speech communities as well. We hope that the ethical challenges outlined here will motivate other researchers working on language technology to also be aware of the differences between languages in ``low-resource settings" but with much larger speaker populations, and indigenous endangered languages; and to be attentive to the needs of the speech communities of the latter. 

\bibliography{anthology,custom}

\begin{thebibliography}{54}
\expandafter\ifx\csname natexlab\endcsname\relax\def\natexlab#1{#1}\fi

\bibitem[{Arppe et~al.(2021)Arppe, Good, Harrigan, Hulden, Lachler, Moeller,
  Palmer, Silfverberg, and Schwartz}]{computel-2021-use}
Antti Arppe, Jeff Good, Atticus Harrigan, Mans Hulden, Jordan Lachler, Sarah
  Moeller, Alexis Palmer, Miikka Silfverberg, and Lane Schwartz, editors. 2021.
\newblock \href {https://aclanthology.org/2021.computel-1.0} {\emph{Proceedings
  of the 4th Workshop on the Use of Computational Methods in the Study of
  Endangered Languages Volume 1 (Papers)}}. Association for Computational
  Linguistics, Online.

\bibitem[{Baevski et~al.(2021)Baevski, Hsu, Conneau, and
  Auli}]{baevski2021unsupervised}
Alexei Baevski, Wei-Ning Hsu, Alexis Conneau, and Michael Auli. 2021.
\newblock Unsupervised speech recognition.
\newblock \emph{arXiv preprint arXiv:2105.11084}.

\bibitem[{Bettinson and Bird(2017)}]{bettinson-bird-2017-developing}
Mat Bettinson and Steven Bird. 2017.
\newblock \href {https://doi.org/10.18653/v1/W17-0121} {Developing a suite of
  mobile applications for collaborative language documentation}.
\newblock In \emph{Proceedings of the 2nd Workshop on the Use of Computational
  Methods in the Study of Endangered Languages}, pages 156--164.

\bibitem[{Bird(2020)}]{bird-2020-decolonising}
Steven Bird. 2020.
\newblock \href {https://doi.org/10.18653/v1/2020.coling-main.313}
  {Decolonising speech and language technology}.
\newblock In \emph{Proceedings of the 28th International Conference on
  Computational Linguistics}, pages 3504--3519, Barcelona, Spain (Online).
  International Committee on Computational Linguistics.

\bibitem[{Bird and Chiang(2012)}]{bird-chiang-2012-machine}
Steven Bird and David Chiang. 2012.
\newblock \href {https://aclanthology.org/C12-2013} {Machine translation for
  language preservation}.
\newblock In \emph{Proceedings of {COLING} 2012: Posters}, pages 125--134,
  Mumbai, India. The COLING 2012 Organizing Committee.

\bibitem[{Blasi et~al.(2021)Blasi, Anastasopoulos, and
  Neubig}]{blasi2021systematic}
Dami{\'a}n Blasi, Antonios Anastasopoulos, and Graham Neubig. 2021.
\newblock Systematic inequalities in language technology performance across the
  world's languages.
\newblock \emph{arXiv preprint arXiv:2110.06733}.

\bibitem[{Cotterell et~al.(2016)Cotterell, Vieira, and
  Sch{\"u}tze}]{cotterell-etal-2016-joint}
Ryan Cotterell, Tim Vieira, and Hinrich Sch{\"u}tze. 2016.
\newblock \href {https://doi.org/10.18653/v1/N16-1080} {A joint model of
  orthography and morphological segmentation}.
\newblock In \emph{Proceedings of the 2016 Conference of the North {A}merican
  Chapter of the Association for Computational Linguistics: Human Language
  Technologies}, pages 664--669, San Diego, California. Association for
  Computational Linguistics.

\bibitem[{Czaykowska-Higgins(2009)}]{czaykowska2009research}
Ewa Czaykowska-Higgins. 2009.
\newblock Research models, community engagement, and linguistic fieldwork:
  Reflections on working within canadian indigenous communities.
\newblock \emph{Language documentation \& conservation}, 3(1):182--215.

\bibitem[{Dixon(1970)}]{dixon1970proto}
Robert Malcolm~Ward Dixon. 1970.
\newblock Proto-australian laminals.
\newblock \emph{Oceanic Linguistics}, pages 79--103.

\bibitem[{Eberhard et~al.(2021)Eberhard, Simons, and
  Fennig}]{campbell2008ethnologue}
David~M. Eberhard, Gary~F. Simons, and Charles~D. Fennig. 2021.
\newblock \emph{Ethnologue: Languages of the World, Twenty-fourth edition}.
\newblock SIL International.

\bibitem[{Eyig{\"o}z et~al.(2013)Eyig{\"o}z, Gildea, and
  Oflazer}]{eyigoz-etal-2013-simultaneous}
Elif Eyig{\"o}z, Daniel Gildea, and Kemal Oflazer. 2013.
\newblock \href {https://www.aclweb.org/anthology/N13-1004} {Simultaneous
  word-morpheme alignment for statistical machine translation}.
\newblock In \emph{Proceedings of the 2013 Conference of the North {A}merican
  Chapter of the Association for Computational Linguistics: Human Language
  Technologies}, pages 32--40, Atlanta, Georgia. Association for Computational
  Linguistics.

\bibitem[{Grenoble(2017)}]{leonard2017}
Lenore~A. Grenoble. 2017.
\newblock Producing language reclamation by decolonising `language'.
\newblock In Wesley Leonard and Haley De~Korne, editors, \emph{The {{Cambridge
  Handbook}} of {{Endangered Languages}}}, volume~14, pages 15--36. London: EL
  Publishing.

\bibitem[{Hale(1983)}]{hale1983warlpiri}
Ken Hale. 1983.
\newblock Warlpiri and the grammar of non-configurational languages.
\newblock \emph{Natural Language \& Linguistic Theory}, 1(1):5--47.

\bibitem[{Hale(1992)}]{hale1992endangered}
Ken Hale. 1992.
\newblock Endangered languages: {O}n endangered languages and the safeguarding
  of diversity.
\newblock \emph{language}, 68(1):1--42.

\bibitem[{Harvey(2015)}]{harveyNativeTonguesColonialism2015}
Sean~P. Harvey. 2015.
\newblock \emph{Native Tongues: Colonialism and Race from Encounter to the
  Reservation}.
\newblock {Harvard University Press}, {Cambridge, Massachusetts}.

\bibitem[{Henrich et~al.(2010)Henrich, Heine, and
  Norenzayan}]{henrich2010weirdest}
Joseph Henrich, Steven~J Heine, and Ara Norenzayan. 2010.
\newblock The weirdest people in the world?
\newblock \emph{Behavioral and brain sciences}, 33(2-3):61--83.

\bibitem[{Hinton(1994)}]{hinton1994flutes}
Leanne Hinton. 1994.
\newblock \emph{Flutes of Fire: Essays on California Indian Languages.}
\newblock ERIC.

\bibitem[{Hinton(2013)}]{hornberger2014bringing}
Leanne Hinton. 2013.
\newblock \emph{Bringing our languages home: Language revitalization for
  families}.
\newblock Heyday Books.

\bibitem[{Jimerson and Prud{'}hommeaux(2018)}]{jimerson-prudhommeaux-2018-asr}
Robbie Jimerson and Emily Prud{'}hommeaux. 2018.
\newblock \href {https://aclanthology.org/L18-1657} {{ASR} for documenting
  acutely under-resourced indigenous languages}.
\newblock In \emph{Proceedings of the Eleventh International Conference on
  Language Resources and Evaluation ({LREC} 2018)}, Miyazaki, Japan. European
  Language Resources Association (ELRA).

\bibitem[{Joshi et~al.(2020)Joshi, Santy, Budhiraja, Bali, and
  Choudhury}]{joshi-etal-2020-state}
Pratik Joshi, Sebastin Santy, Amar Budhiraja, Kalika Bali, and Monojit
  Choudhury. 2020.
\newblock \href {https://doi.org/10.18653/v1/2020.acl-main.560} {The state and
  fate of linguistic diversity and inclusion in the {NLP} world}.
\newblock In \emph{Proceedings of the 58th Annual Meeting of the Association
  for Computational Linguistics}, pages 6282--6293.

\bibitem[{Kann et~al.(2018)Kann, Mager~Hois, Meza-Ruiz, and
  Sch{\"u}tze}]{kann-etal-2018-fortification}
Katharina Kann, Jesus~Manuel Mager~Hois, Ivan~Vladimir Meza-Ruiz, and Hinrich
  Sch{\"u}tze. 2018.
\newblock \href {https://doi.org/10.18653/v1/N18-1005} {Fortification of neural
  morphological segmentation models for polysynthetic minimal-resource
  languages}.
\newblock In \emph{Proceedings of the 2018 Conference of the North {A}merican
  Chapter of the Association for Computational Linguistics: Human Language
  Technologies, Volume 1 (Long Papers)}, pages 47--57.

\bibitem[{Kazantseva et~al.(2018)Kazantseva, Maracle, Maracle, and
  Pine}]{kazantseva-etal-2018-kawennon}
Anna Kazantseva, Owennatekha~Brian Maracle, Ronkwe{'}tiy{\'o}hstha~Josiah
  Maracle, and Aidan Pine. 2018.
\newblock \href {https://aclanthology.org/W18-4806} {{K}awenn{\'o}n:nis: the
  {W}ordmaker for {K}anyen{'}k{\'e}ha}.
\newblock In \emph{Proceedings of the Workshop on Computational Modeling of
  Polysynthetic Languages}, pages 53--64, Santa Fe, New Mexico, USA.
  Association for Computational Linguistics.

\bibitem[{Khare et~al.(2021)Khare, Mittal, Diwan, Sarawagi, Jyothi, and
  Bharadwaj}]{khare2021low}
Shreya Khare, Ashish Mittal, Anuj Diwan, Sunita Sarawagi, Preethi Jyothi, and
  Samarth Bharadwaj. 2021.
\newblock Low resource {ASR}: The surprising effectiveness of high resource
  transliteration.
\newblock In \emph{The Annual Conference of the International Speech
  Communication Association (Interspeech)}, pages 1529--1533.

\bibitem[{Kuhn et~al.(2020)Kuhn, Davis, D{\'e}silets, Joanis, Kazantseva,
  Knowles, Littell, Lothian, Pine, Running~Wolf, Santos, Stewart, Boulianne,
  Gupta, Maracle~Owennat{\'e}kha, Martin, Cox, Junker, Sammons, Torkornoo,
  Thanyeht{\'e}nhas~Brinklow, Child, Farley, Huggins-Daines, Rosenblum, and
  Souter}]{kuhn-etal-2020-indigenous}
Roland Kuhn, Fineen Davis, Alain D{\'e}silets, Eric Joanis, Anna Kazantseva,
  Rebecca Knowles, Patrick Littell, Delaney Lothian, Aidan Pine, Caroline
  Running~Wolf, Eddie Santos, Darlene Stewart, Gilles Boulianne, Vishwa Gupta,
  Brian Maracle~Owennat{\'e}kha, Akwirat{\'e}kha{'} Martin, Christopher Cox,
  Marie-Odile Junker, Olivia Sammons, Delasie Torkornoo, Nathan
  Thanyeht{\'e}nhas~Brinklow, Sara Child, Beno{\^\i}t Farley, David
  Huggins-Daines, Daisy Rosenblum, and Heather Souter. 2020.
\newblock \href {https://doi.org/10.18653/v1/2020.coling-main.516} {The
  {I}ndigenous {L}anguages {T}echnology project at {NRC} {C}anada: {A}n
  empowerment-oriented approach to developing language software}.
\newblock In \emph{Proceedings of the 28th International Conference on
  Computational Linguistics}, pages 5866--5878, Barcelona, Spain (Online).
  International Committee on Computational Linguistics.

\bibitem[{Lachler et~al.(2018)Lachler, Antonsen, Trosterud, Moshagen, and
  Arppe}]{lachler-etal-2018-modeling}
Jordan Lachler, Lene Antonsen, Trond Trosterud, Sjur Moshagen, and Antti Arppe.
  2018.
\newblock \href {https://aclanthology.org/L18-1368} {Modeling {N}orthern
  {H}aida verb morphology}.
\newblock In \emph{Proceedings of the Eleventh International Conference on
  Language Resources and Evaluation ({LREC} 2018)}, Miyazaki, Japan. European
  Language Resources Association (ELRA).

\bibitem[{Lane and Bird(2020)}]{lane-bird-2020-bootstrapping}
William Lane and Steven Bird. 2020.
\newblock \href {https://doi.org/10.18653/v1/2020.acl-main.594} {Bootstrapping
  techniques for polysynthetic morphological analysis}.
\newblock In \emph{Proceedings of the 58th Annual Meeting of the Association
  for Computational Linguistics}, pages 6652--6661.

\bibitem[{Leonard(2011)}]{leonard2011challenging}
Wesley Leonard. 2011.
\newblock Challenging ``extinction" through modern {M}iami language practices.
\newblock \emph{American Indian Culture and Research Journal}, 35(2):135--160.

\bibitem[{Lewis et~al.(2015)Lewis, Simons, and Fennig}]{lewis2015ethnologue}
M~Paul Lewis, Gary~F Simons, and Charles~D Fennig. 2015.
\newblock Ethnologue: languages of {E}cuador.
\newblock \emph{Ethnologue}.

\bibitem[{Liu et~al.(2021)Liu, Jimerson, and
  Prud{'}hommeaux}]{liu-etal-2021-morphological}
Zoey Liu, Robert Jimerson, and Emily Prud{'}hommeaux. 2021.
\newblock \href {https://doi.org/10.18653/v1/2021.americasnlp-1.10}
  {Morphological segmentation for {S}eneca}.
\newblock In \emph{Proceedings of the First Workshop on Natural Language
  Processing for Indigenous Languages of the Americas}, pages 90--101.

\bibitem[{Mager et~al.(2020)Mager, {\c{C}}etino{\u{g}}lu, and
  Kann}]{mager-etal-2020-tackling}
Manuel Mager, {\"O}zlem {\c{C}}etino{\u{g}}lu, and Katharina Kann. 2020.
\newblock \href {https://doi.org/10.18653/v1/2020.emnlp-main.423} {Tackling the
  low-resource challenge for canonical segmentation}.
\newblock In \emph{Proceedings of the 2020 Conference on Empirical Methods in
  Natural Language Processing (EMNLP)}, pages 5237--5250.

\bibitem[{Mager et~al.(2021)Mager, Oncevay, Rios, Ruiz, Palmer, Neubig, and
  Kann}]{americasnlp-2021-natural}
Manuel Mager, Arturo Oncevay, Annette Rios, Ivan Vladimir~Meza Ruiz, Alexis
  Palmer, Graham Neubig, and Katharina Kann, editors. 2021.
\newblock \href {https://aclanthology.org/2021.americasnlp-1.0}
  {\emph{Proceedings of the First Workshop on Natural Language Processing for
  Indigenous Languages of the Americas}}. Association for Computational
  Linguistics, Online.

\bibitem[{Meek(2012)}]{meek2012we}
Barbra~A Meek. 2012.
\newblock \emph{We are our language: An ethnography of language revitalization
  in a Northern Athabaskan community}.
\newblock University of Arizona Press.

\bibitem[{Mitra et~al.(2016)Mitra, Kathol, Amith, and
  Garc{\'\i}a}]{mitra2016automatic}
Vikramjit Mitra, Andreas Kathol, Jonathan~D Amith, and Rey~Castillo
  Garc{\'\i}a. 2016.
\newblock Automatic speech transcription for low-resource languages-the case of
  yolox{\'o}chitl mixtec (mexico).
\newblock In \emph{INTERSPEECH}, pages 3076--3080.

\bibitem[{Morris et~al.(2021)Morris, Jimerson, and Prud'hommeaux}]{ethan2021}
Ethan Morris, Robert Jimerson, and Emily Prud'hommeaux. 2021.
\newblock One size does not fit all in resource-constrained {ASR}.
\newblock In \emph{Proceedings of Interspeech}, pages 4354--4358.

\bibitem[{Povey et~al.(2011)Povey, Ghoshal, Boulianne, Burget, Glembek, Goel,
  Hannemann, Motlicek, Qian, Schwarz et~al.}]{povey2011kaldi}
Daniel Povey, Arnab Ghoshal, Gilles Boulianne, Lukas Burget, Ondrej Glembek,
  Nagendra Goel, Mirko Hannemann, Petr Motlicek, Yanmin Qian, Petr Schwarz,
  et~al. 2011.
\newblock The {K}aldi speech recognition toolkit.
\newblock In \emph{IEEE 2011 workshop on automatic speech recognition and
  understanding}, CONF. IEEE Signal Processing Society.

\bibitem[{Prud'hommeaux et~al.(2021)Prud'hommeaux, Jimerson, Hatcher, and
  Michelson}]{prudhommeaux2021}
Emily Prud'hommeaux, Robbie Jimerson, Richard Hatcher, and Karin Michelson.
  2021.
\newblock Automatic speech recognition for supporting endangered language
  documentation.
\newblock \emph{Language Documentation and Conservation}.

\bibitem[{Richardson(2018)}]{crystal2018}
Crystal Richardson. 2018.
\newblock Uhyanavararatih: {A} {C}all {A}cross {T}he {D}ivide.
\newblock Master's thesis, University of California, Davis.

\bibitem[{Richardson and Brucell(1993)}]{richardson1993now}
Nancy Richardson and Suzanne Brucell. 1993.
\newblock \emph{Now you’re speaking—Karuk}.
\newblock Center for Indian Community Development.

\bibitem[{Rigney(1999)}]{rigney1999internationalization}
Lester-Irabinna Rigney. 1999.
\newblock Internationalization of an indigenous anticolonial cultural critique
  of research methodologies: A guide to indigenist research methodology and its
  principles.
\newblock \emph{Wicazo sa review}, 14(2):109--121.

\bibitem[{Rigney(2006)}]{rigney2006indigenous}
Lester-Irabinna Rigney. 2006.
\newblock Indigenous australian views on knowledge production and indigenist
  research.
\newblock \emph{Indigenous peoples’ wisdom and power}, pages 32--49.

\bibitem[{Rood and Taylor(1996)}]{rood1996sketch}
David~S Rood and Allan~R Taylor. 1996.
\newblock Sketch of {L}akhota, a {S}iouan language.
\newblock \emph{Handbook of North American Indians}, 17:440--482.

\bibitem[{Ross et~al.(2002)Ross, Everett, Jelinek, Harley, Perkins, Willie,
  Grinevald, and Austin}]{remember}
John~Robert Ross, Daniel~E Everett, Eloise Jelinek, Heidi Harley, Ellavina
  Perkins, MaryAnn Willie, Colette Grinevald, and Peter~K Austin. 2002.
\newblock Remembering {K}enneth {L}. {H}ale.
\newblock \emph{Linguistic Typology}, (6):137--153.

\bibitem[{Ryon(2002)}]{ryon2002cajun}
Dominique Ryon. 2002.
\newblock Cajun french, sociolinguistic knowledge, and language loss in
  louisiana.
\newblock \emph{Journal of Language, Identity, and Education}, 1(4):279--293.

\bibitem[{Seeker and {\c{C}}etino{\u{g}}lu(2015)}]{seeker-cetinoglu-2015-graph}
Wolfgang Seeker and {\"O}zlem {\c{C}}etino{\u{g}}lu. 2015.
\newblock \href {https://doi.org/10.1162/tacl_a_00144} {A graph-based lattice
  dependency parser for joint morphological segmentation and syntactic
  analysis}.
\newblock \emph{Transactions of the Association for Computational Linguistics},
  3:359--373.

\bibitem[{Shi et~al.(2021{\natexlab{a}})Shi, Amith, Castillo~Garc{\'\i}a,
  Guadalupe~Sierra, Duh, and Watanabe}]{shi-etal-2021-leveraging}
Jiatong Shi, Jonathan~D. Amith, Rey Castillo~Garc{\'\i}a, Esteban
  Guadalupe~Sierra, Kevin Duh, and Shinji Watanabe. 2021{\natexlab{a}}.
\newblock \href {https://aclanthology.org/2021.eacl-main.96} {Leveraging
  end-to-end {ASR} for endangered language documentation: An empirical study on
  {Y}olox{\'o}chitl {M}ixtec}.
\newblock In \emph{Proceedings of the 16th Conference of the European Chapter
  of the Association for Computational Linguistics: Main Volume}, pages
  1134--1145.

\bibitem[{Shi et~al.(2021{\natexlab{b}})Shi, Amith, Chang, Dalmia, Yan, and
  Watanabe}]{shi-etal-2021-highland}
Jiatong Shi, Jonathan~D. Amith, Xuankai Chang, Siddharth Dalmia, Brian Yan, and
  Shinji Watanabe. 2021{\natexlab{b}}.
\newblock \href {https://doi.org/10.18653/v1/2021.americasnlp-1.7} {{H}ighland
  {P}uebla {N}ahuatl speech translation corpus for endangered language
  documentation}.
\newblock In \emph{Proceedings of the First Workshop on Natural Language
  Processing for Indigenous Languages of the Americas}, pages 53--63.

\bibitem[{Skilton(2017)}]{skilton2017three}
Amalia Skilton. 2017.
\newblock \emph{Three speakers, four dialects: Documenting variation in an
  endangered Amazonian language}.
\newblock University of Hawai'i Press.

\bibitem[{Skutnabb-Kangas(2000)}]{skutnabb2000linguistic}
Tove Skutnabb-Kangas. 2000.
\newblock \emph{Linguistic genocide in education--or worldwide diversity and
  human rights?}
\newblock Routledge.

\bibitem[{Thai et~al.(2020)Thai, Jimerson, Ptucha, and
  Prud’hommeaux}]{thai2020fully}
Bao Thai, Robert Jimerson, Raymond Ptucha, and Emily Prud’hommeaux. 2020.
\newblock Fully convolutional asr for less-resourced endangered languages.
\newblock In \emph{Proceedings of the 1st Joint Workshop on Spoken Language
  Technologies for Under-resourced languages (SLTU) and Collaboration and
  Computing for Under-Resourced Languages (CCURL)}, pages 126--130.

\bibitem[{Watanabe et~al.(2018)Watanabe, Hori, Karita, Hayashi, Nishitoba,
  Unno, {Enrique Yalta Soplin}, Heymann, Wiesner, Chen, Renduchintala, and
  Ochiai}]{watanabe2018espnet}
Shinji Watanabe, Takaaki Hori, Shigeki Karita, Tomoki Hayashi, Jiro Nishitoba,
  Yuya Unno, Nelson {Enrique Yalta Soplin}, Jahn Heymann, Matthew Wiesner,
  Nanxin Chen, Adithya Renduchintala, and Tsubasa Ochiai. 2018.
\newblock \href {https://doi.org/10.21437/Interspeech.2018-1456} {{ESPnet}:
  {E}nd-to-{E}nd {S}peech {P}rocessing {T}oolkit}.
\newblock In \emph{Proceedings of Interspeech}, pages 2207--2211.

\bibitem[{Whalen et~al.(2016)Whalen, Moss, and Baldwin}]{whalen2016healing}
Douglas~H Whalen, Margaret Moss, and Daryl Baldwin. 2016.
\newblock Healing through language: Positive physical health effects of
  indigenous language use.
\newblock \emph{F1000Research}, 5(852):852.

\bibitem[{Whaley(2011)}]{whaley2011some}
Lindsay~J Whaley. 2011.
\newblock Some ways to endanger an endangered language project.
\newblock \emph{Language and Education}, 25(4):339--348.

\bibitem[{Zahrer et~al.(2020)Zahrer, Zgank, and
  Schuppler}]{zahrer-etal-2020-towards}
Alexander Zahrer, Andrej Zgank, and Barbara Schuppler. 2020.
\newblock \href {https://aclanthology.org/2020.lrec-1.353} {Towards building an
  automatic transcription system for language documentation: Experiences from
  {M}uyu}.
\newblock In \emph{Proceedings of the 12th Language Resources and Evaluation
  Conference}, pages 2893--2900.

\bibitem[{Zhang et~al.(2020)Zhang, Frey, and Bansal}]{zhang-etal-2020-chren}
Shiyue Zhang, Benjamin Frey, and Mohit Bansal. 2020.
\newblock \href {https://doi.org/10.18653/v1/2020.emnlp-main.43} {{C}hr{E}n:
  {C}herokee-{E}nglish machine translation for endangered language
  revitalization}.
\newblock In \emph{Proceedings of the 2020 Conference on Empirical Methods in
  Natural Language Processing (EMNLP)}, pages 577--595.

\end{thebibliography}
\bibliographystyle{acl_natbib}



\end{document}